\documentclass[11pt,a4paper]{article}
\usepackage[hyperref]{emnlp2020}
\usepackage{times}
\usepackage{latexsym}

\usepackage{microtype}
\usepackage{graphicx}
\usepackage{multirow}
\usepackage{booktabs}
\usepackage{mathtools}
\usepackage{cuted}
\usepackage{subcaption}

\usepackage{enumitem}% http://ctan.org/pkg/enumitem
\usepackage{breqn}

\usepackage{amsmath}
\usepackage{tikz}
\usepackage{placeins}
\usepackage{amsthm}
\usepackage{array}

\usepackage{subfiles} % Best loaded last in the preamble

\aclfinalcopy % Uncomment this line for the final submission
\newcolumntype{P}[1]{>{\centering\arraybackslash}p{#1}}

\theoremstyle{definition}

\DeclareMathOperator*{\argmax}{arg\,max}
\DeclarePairedDelimiter\ceil{\lceil}{\rceil}

\title{ExplanationLP: Abductive Reasoning for Explainable Science Question Answering}

\author{
  Mokanarangan Thayaparan, Marco Valentino, Andr\'e Freitas \\
  Department of Computer Science\\
  University of Manchester \\
  \texttt{\{mokanarangan.thayaparan, marco.valentino, andre.freitas\}}\\ \texttt{@manchester.ac.uk}
  \\}

\date{}

\begin{document}
\maketitle

\begin{abstract}
We propose a novel approach for answering and explaining multiple-choice science questions by reasoning on grounding and abstract inference chains. This paper frames question answering as an abductive reasoning problem, constructing \textit{plausible} explanations for each choice and then selecting the candidate with the \textit{best} explanation as the final answer. Our system, \textit{ExplanationLP}, elicits explanations by constructing a weighted graph of relevant facts for each candidate answer and extracting the facts that satisfy certain structural and semantic constraints. To extract the explanations, we employ a linear programming formalism designed to select the optimal subgraph. The graphs' weighting function is composed of a set of parameters, which we fine-tune to optimize answer selection performance. We carry out our experiments on the WorldTree and ARC-Challenge corpus to empirically demonstrate the following conclusions: (1) Grounding-Abstract inference chains provides the semantic control to perform explainable abductive reasoning (2) Efficiency and robustness in learning with a fewer number of parameters by outperforming contemporary explainable and transformer-based approaches in a similar setting (3) Generalisability by outperforming SOTA explainable approaches on general science question sets.
\end{abstract}
\section{Introduction}

Science Question Answering (QA) remains a fundamental challenge in Natural Language Processing and AI, as it requires complex forms of inference including causal, model-based and example-based reasoning~\cite{jansen2018multi,clark2018think,jansen2016s,clark2013study}. Current state-of-the-art (SOTA) approaches for Science QA are dominated by transformer-based models~\cite{devlin2019bert,zhu2019freelb,Sun_2019}. However, despite remarkable performances on answer prediction, these approaches are black-box by nature, lacking of the capability of providing \textit{explanations} for their predictions~\cite{miller2019explanation,biran2017explanation,jansen2016s}.

\textit{Explainable Science QA} (XSQA) is often framed as an \textit{abductive reasoning} problem~\cite{peirce1960collected}. Abductive reasoning represents a distinct inference process, known as \emph{inference to the best explanation} \cite{lipton2004inference}, which starts from a set of complete or incomplete observations to find the hypothesis, from a set of plausible alternatives, that \emph{best} explains the observations. Several approaches~\cite{semanticilp2018aaai,jansen2017framing,khot2017answering,khashabi2016question} employ this form of reasoning for multiple-choice science questions to build a set of plausible explanations for each candidate answer and select the one with the best explanation as the final answer. 

XSQA solvers~\cite{semanticilp2018aaai,khot2017answering,khashabi2016question} typically treat question answering as a multi-hop graph traversal problem. Here, the solver attempts to compose multiple facts that connect the question to a candidate answer. These \textit{multi-hop} approaches have shown diminishing returns with an increasing number of hops~\cite{jansen2018worldtree,jansen2018multi}. ~\citet{fried2015higher} conclude that this phenomenon is due to \textit{semantic drift} -- i.e., as the number of aggregated facts increases, so does the probability of drifting out of context. ~\citet{khashabi2019capabilities} propose a theoretical framework, empirically supported by~\citet{jansen2018worldtree,fried2015higher}, attesting that ongoing efforts with \textit{very long} multi-hop reasoning are unlikely to succeed, emphasizing the need for a \textit{richer} representation with fewer hops and higher importance to abstraction and grounding mechanisms.

% \begin{figure}[t]
%     \centering
%     \small 
%     \includegraphics[width=\columnwidth]{images/unification.pdf}
%     \caption{An example extracted depicting a question, answer and gold explanatory facts supporting the answer divided into grounding and abstract facts. Texts sharing the same highlight color indicates lexical overlap.}
%     \label{fig:grounding_central}
% \end{figure}

\begin{figure*}[t!]
    \centering
    \includegraphics[width=\textwidth]{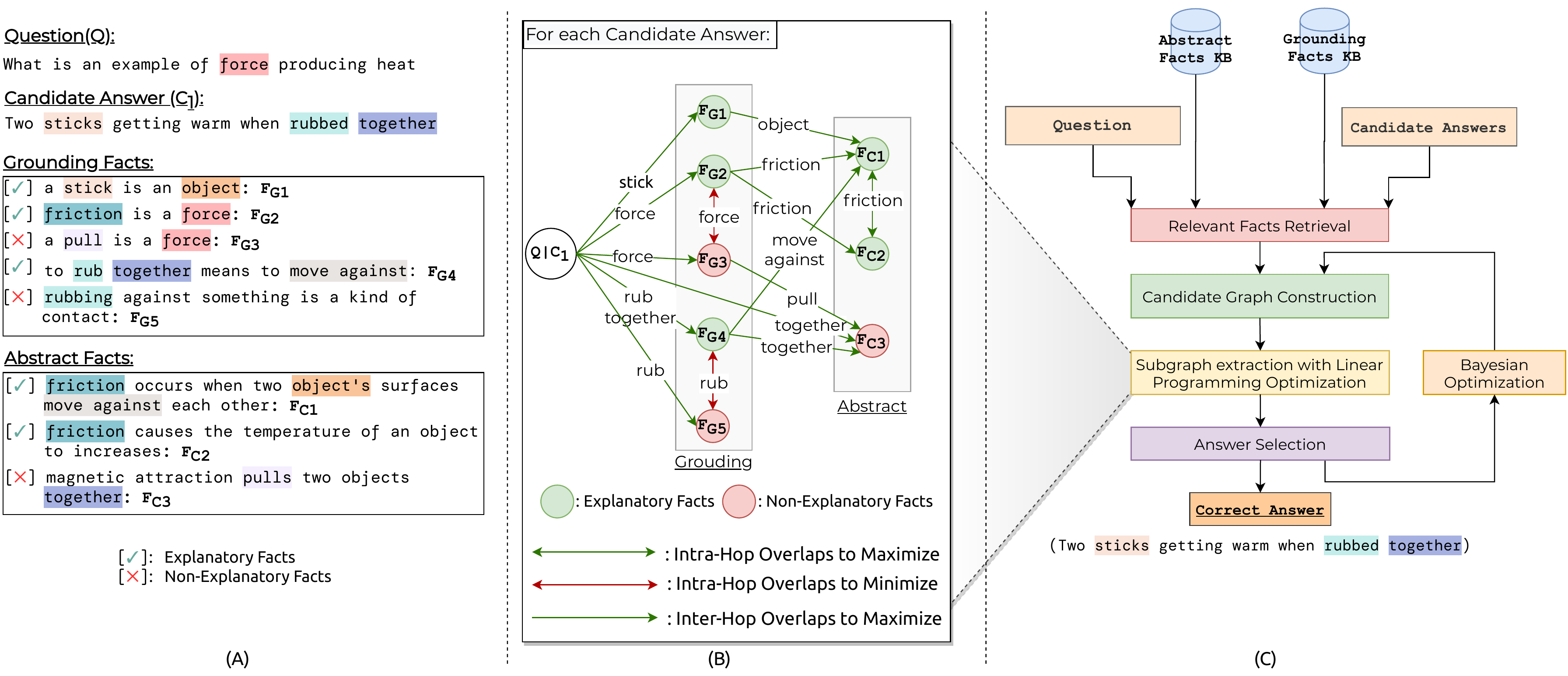}
    \caption{Overview of our approach: (A) Depicts a question, answer and set of facts retrieved from a fact retrieval approach. (B) Illustrates the optimization process behind extracting explanatory facts for the provided question, candidate answer and facts. (C) Details the end-to-end architecture diagram.  }
    \label{fig:end_end}
\end{figure*}

Consider the example in Figure~\ref{fig:end_end}A, where the central concept the question examines is the understanding of \textit{friction}. Here, the challenge for a QA solver is to identify the core scientific facts (\textit{Abstract Facts}) that best explain the answer. To achieve this goal, a QA solver should be able first to go from \textit{force} to \textit{friction}, \textit{stick} to \textit{object} and \textit{rubbing together} to \textit{move against}. These are the \textit{Grounding Facts} that link generic or abstract concepts in a core scientific statement to specific terms occurring in question and candidate answer~\cite{jansen2018worldtree}. The grounding process is followed by the identification of the abstract facts about \textit{friction}. Even though a complete explanation for this question would require the composition of five facts, to successfully derive the correct answer it is possible to reduce the global reasoning in two hops, modeling it with grounding and abstract facts.  

In line with these observations, this work presents a novel approach that explicitly models abstract and grounding mechanisms by grouping explanatory facts into \textit{grounding} and \textit{abstract}. These facts are then used to perform abductive reasoning via linear programming combined with Bayesian optimization. We demonstrate that this paradigm can be used to answer and explain complex science questions with fewer learning parameters. In summary, the contributions of the paper are: 
\\
\noindent\textbf{(1)} Demonstrate how grounding-abstract chains facilitates semantic control for explainable abductive reasoning\\
\noindent\textbf{(2)} Exhibit how Bayesian optimization with linear programming can be employed with few learning parameters and achieve better performance and robustness when compared to transformer-based and graph-based multihop reasoning approaches.\\
\textbf{(3)}  Show how ExplanationLP can outperform SOTA explainable approaches for science question answering.

% \begin{enumerate}[leftmargin=*,itemsep=1em]
%     \itemsep0em 
%     \item An Explainable Question Answering approach that outperforms transformer-based approaches for science questions (See Section~\ref{sec:ans_selection},~\ref{sec:explanation_retrieval},~\ref{sec:error_analysis})
%     % \item Modeling grounding-abstract inference chains using linear programming for question answering (See Section~\ref{sec:ablation})
%     \item An empirical evaluation highlighting the impact of grounding-abstract inference chains for answer selection via ablation experiments (See Section~\ref{sec:ablation})
%     \item Employing Bayesian Optimization with linear programming for explainable question answering (See Section~\ref{sec:ablation})
% \end{enumerate}

\section{ExplanationLP: Abductive Reasoning with Linear Programming}
\label{sec:approach}

A multiple-choice science question $Q$ requires selecting the correct answer $c_{ans}$ from a set of candidate choices $C = \{c_1, c_2, \dots c_n\}$. ExplanationLP answers and explains multiple-choice science questions via abductive reasoning. Specifically, the task of answering multiple-choice science questions is reformulated as the problem of finding the candidate answer that is supported by the best explanation. For each candidate answer $c_i \in C$, ExplanationLP attempts to construct a plausible explanation, quantifying its quality through scoring functions rooted in the semantic relevance of facts while preserving structural constraints imposed via grounding-abstract chains. The model operates through the following steps (See Figure~\ref{fig:end_end}C):

\noindent \textbf{(1) Relevant facts retrieval}:
    Given a question ($Q$) and candidate answers $C=\{c_1,~c_2,~c_3, ...,c_n\}$, for each candidate answer $c_i$ we query the knowledge bases  using a fact retrieval approach (e.g: BM25, Unification Retrieval~\cite{valentino2020unification}) to retrieve the top $k$ relevant facts $F^{c_i}$. We achieve the retrieval by concatenating question and candidate  answer ($Q||c_i$) to retrieve the top $l$ relevant \textit{grounding} facts $F^{c_i}_{G} = \{f^{c_i}_{1}, f^{c_i}_{2}, f^{c_i}_{3}, ..., f^{c_i}_{l}\}$ from a knowledge base containing grounding facts (\textit{Grounding Facts KB}) and  top $m$ relevant \textit{abstract} facts $F^{c_i}_{A} = \{f^{c_i}_{1}, f^{c_i}_{2}, f^{c_i}_{3}, ..., f^{c_i}_{m}\}$ from a knowledge base containing abstract facts (\textit{Abstract Facts KB}) such that $F^{c_i}=F^{c_i}_{A} \cup F^{c_i}_{G}  $ and $l + m = k$

\noindent \textbf{(2) Candidate graph construction}: For each candidate answer $c_i$ we build a weighted undirected graph $G^{c_i} = (V^{c_i},~E^{c_i},~\omega_v,~\omega_e)$ with vertices $V^{c_i} \in \{ \{Q||c_i\} \cup F^{c_i}\}$, edges $E^{c_i}=\{ (v^{c_i}_j,v^{c_i}_k) : v^{c_i}_j, v^{c_i}_k \in V^{c_i},~v^{c_i}_j \ne v^{c_i}_k,  \text{has-lexical-overlap}(v_i^{c_i},v^{c_i}_j)\}$, edge-weight function $\omega_e(e_i;\theta)$ and node-weight function $\omega_v(v_i;\theta)$ where $e_i \in E^{c_i}$, $v_i \in V^{c_i}$ and $\theta \in {\rm I\!R^n}$ is a learnable parameter.
        
\noindent\textbf{(3) Subgraph extraction with linear programming optimization}: 
    For each graph $G^{c_i}$, we apply linear programming optimization with objective function $\Omega(G^{c_i};\theta)$ to obtain the optimal subgraph $\hat{G}^{c_i}=(\hat{V}^{c_i},~\hat{E}^{c_i})$. The linear programming constraints are designed to emulate the abductive reasoning over grounding-abstract inference chains. During the training phase, we update the weight functions $\omega_v$ and $\omega_e$  by tuning the parameters ($\theta$) to optimize for answer selection using Bayesian Optimization.
        
\noindent \textbf{(4) Answer Selection}:  We apply steps 1-3 for each candidate answer $C=\{c_1,~c_2,~c_3, ...,~c_n\}$  to obtain explanation graphs $G^Q=\{\hat{G}^{c_1},~\hat{G}^{c_2},~\hat{G}^{c_3},...,~\hat{G}^{c_n}\}$. The candidate answer with the highest value for $\Omega$ is selected as the final answer.

\subsection{Constructing Candidate Answer Graph}
\label{sec:construct_graph}

% \begin{figure}
%     \centering
%     \small
%     \includegraphics[width=\columnwidth]{images/approach_expl.pdf}
%     \caption{The example depicts a question, answer and set of explanatory facts retrieved from an information retrieval approach.}
%     \label{fig:information_retreival}
% \end{figure}

In order to answer a question we need to retrieve semantically relevant but still diverse explanations for each candidate answer. In the example from Figure~\ref{fig:end_end}A, $F_{G1}$ and $F_{G5}$ contain relevant information to answer the question. However, only  $F_{G1}$ will lead to the correct abstract fact, rendering $F_{G5}$ as redundant. On the other hand, $F_{G4}$ is not relevant to the question and selecting it would lead to a spurious inference chain with $F_{C3}$. As visualized in Figure~\ref{fig:end_end}C, our approach imposes several structural and semantic constraints to tackle this challenge. Grounding-Abstract chains play a crucial role in the solution. To achieve this, we use the following scores:

% \begin{itemize}[leftmargin=*]
     \noindent\textbf{Relevance ($R$)}:  We use the relevance score obtained from the upstream retrieval approach to represent each node's weights in $G^{c_i}$.
     
    %  This score aims to reward the selection of facts with higher relevance, limiting the aggregation of irrelevant facts.\\
     \noindent \textbf{Overlap ($O$)}: Overlap score represents the lexical overlap of the unique terms between facts. The unique terms in a fact, after being lemmatized and stripped of stopwords, are the verbs and nouns also existing in WordNet. Unique terms from $f^{c_i}_i$ are represented by $t(f^{c_i}_i)$. We use the overlap scores to represent the edge weights between nodes. The overlap scoring function is as follows:
    % \begin{itemize}[leftmargin=*]
    \itemsep0em 
        % \item \textbf{Intra-Hop Overlap ($O_{intra}$)}: Given two facts $f^{c_i}_j$ and $f^{c_i}_k$, $O_{intra}$ score is defined as:
    \begin{equation*}
        % \small
        O(f^{c_i}_j,f^{c_i}_k) = \frac{\vert t(f^{c_i}_j) \cap t(f^{c_i}_k) \vert}{max(\vert t(f^{c_i}_j)\vert, \vert t(f^{c_i}_k) \vert)}
    \end{equation*}

% We use facts with higher relevance score, as it promotes high conceptual alignment with the question and answer. Overlap score is applied to represent all the edges. In the case of grounding-grounding facts, the aim is to minimize the overlap, thus incentivizing the solver to address different parts of the question. Meanwhile, by maximizing overlap between abstract facts, we obtain abstract facts referring to similar topics, avoiding the construction of contradictory or spurious explanations. On the other hand, for question-grounding and grounding-abstract edges, we aim to maximize the overlap to encourage high contextual alignment, so that the inference chains do not drift out of context. 

% By minimizing overlaps between grounding facts, we encourage the solver to address different parts of the question. Meanwhile, by maximizing overlap between abstract facts, we obtain abstract facts referring to similar topics, avoiding the construction of contradictory or spurious explanations.  We also  encourage a high degree of overlaps between multiple hops, preventing the drifting away of chains from the original context. 

We base the construction of the explanatory graphs on the following design principles:
\begin{itemize}[leftmargin=*]
    \item Explanations for science questions can be constructed by multi-hop grounding-abstract chains.
    \item Encouraging facts with higher relevance score limits semantic drift, as it promotes high conceptual alignment with the question and answer. 
    \item Minimizing overlaps in the grounding facts reduces semantic drift, as it promotes a higher coverage of the concepts in the question.
    \item Maximizing overlaps between abstract facts reduces semantic drift, as it forces the abstract facts to refer to similar topics, avoiding the construction of contradictory or spurious explanations.
    \item Encouraging a high degree of overlaps between hops prevent the chains to drift away from the original context.
\end{itemize}

\subsection{Subgraph Selection with Linear Programming (LP)}

In contrast to other LP-based question-answering approaches, which frame the selection of explanations as a connectivity problem, we treat as a rooted maximum-weight connected subgraph problem with a maximum number of $K$ vertices (R-MWCS$_{K}$). This formalism is derived from generalized maximum-weight connected subgraph problem~\cite{loboda2016solving}. R-MWCS$_{K}$ has two parts: objective function to be maximized and constraints to build a connected subgraph. The formal definition of the objective function is as follows: \\

\noindent\textbf{Definition 1.} Given a connected undirected graph $G= (V,E)$ with edge-weight function $\omega_e: E \rightarrow {\rm I\!R}$, node-weight function $\omega_v: V \rightarrow {\rm I\!R}$ , root vertex $r \in V$ and expected number of vertices $K$, the rooted maximum-weight connected subgraph problem with $K$ number of vertices (R-MWCS$_{K}$) problem is finding the connected subgraph $\hat{G}=(\hat{V},\hat{E})$ such that $r \in \hat{V}$,~$\vert V \vert \leq K$ and
\begin{multline*}
    \small
    \Omega(\hat{G};\theta) = 
    \theta_{vw}\sum_{v\in\hat{V}} \omega_v(v;\theta) \\ + \theta_{ew}\sum_{e \in \hat{E}} \omega_e(e;\theta) \rightarrow max
\end{multline*}
\label{def:rmcs}

\begin{table*}[t]
    \centering
   \resizebox{\textwidth}{!}{%
   \begin{tabular}{p{2.8in}m{4.8in}}
    \toprule
    \parbox{2.5in}{\small\begin{align}
    \label{eq:constraint_question}
     y_{v_i} =&~1 &&\text{ if } v_i = Q\\
    \label{eq:constraint_neigbors}
     y_{v_i} \leq&~\sum_{j}y_{v_j} && \forall v_j \in  N_{G^{c_i}}(v_i) \\
     \label{eq:constraint_edge_1}
     z_{v_i,v_j} \leq&~y_{v_i}&& \forall e_{(v_i,v_j)} \in E  \\
     \label{eq:constraint_edge_2}
    z_{v_i,v_j} \leq&~y_{v_j}&& \forall e_{(v_i,v_j)} \in E \\
     \label{eq:constraint_edge_3}
    z_{v_i,v_j} \geq&~y_{v_i}+y_{v_j}-1 && \forall e_{(v_i,v_j)} \in E
\end{align}%
}    & \textbf{Chaining constraint}: Equation~\ref{eq:constraint_question} states that the subgraph should always contain the question node. Inequality~\ref{eq:constraint_neigbors} states that if a vertex is to be part of the subgraph, then at least one of its neighbors with a lexical overlap should also be part of the subgraph. Equation~\ref{eq:constraint_question} and Inequality~\ref{eq:constraint_neigbors} restrict the LP system to construct explanations that originate from the question and perform multi-hop aggregation based on the existence of lexical overlap. Inequalities~\ref{eq:constraint_edge_1},~\ref{eq:constraint_edge_2} and~\ref{eq:constraint_edge_3} state that if two vertices are in the subgraph then the edges connecting the vertices should be also in the subgraph. These inequality constraints will force the LP system to avoid grounding nodes with high overlap regardless of their relevance. 
% Thus incentivizing the solver to address different parts of the question. On the other hand, for abstract-abstract facts, the system would maximize the overlap so that abstract facts do not drift out of context. 
\\
\midrule
\parbox{2.5in}{\begin{align}
    \label{eq:constraint_limit}
     \sum_{i} y_{v_i} \leq&~K &&\forall v_i \in F^{c_i}_{A}
\end{align}%
} &
\textbf{Abstract fact limit constraint}: Equation~\ref{eq:constraint_limit} limits the total number of abstract facts to $K$. By limiting the abstract facts, we dictate the need for grounding facts based on the number of terms present in the question and in the abstract facts.\\
\midrule
\parbox{2.5in}{\small\begin{align}
    \label{eq:constraint_grounding}
     \sum_{v_j}y_{v_i}-2 \geq&-2(1-y_{v_j}) &&\begin{aligned}[t] 
     \forall v_i \in N_{G^{c_i}}(v_j),\\ v_i \in \{F^{c_i}_{A} \cup Q \}, \\ v_j \in F^{c_i}_{G} 
     \end{aligned}
\end{align}
}&
\textbf{Grounding neighbor constraint}: Inequality~\ref{eq:constraint_grounding} states that if a grounding fact is selected, then at least two of its neighbors should be either both abstract facts or a question and an abstract fact. This constraint ensures that grounding facts play the linking role connecting question-abstract facts.\\
    \bottomrule
    \end{tabular}}
    \caption{\small Linear programming constraints employed by ExplanationLP to emulate grounding-abstract inference chains and  extract the optimal subgraph}
    \label{tab:lp_constraints}
\end{table*}

% The weight function $\omega$ is composed of parameters ($\alpha_i$) which we fine-tune using Bayesian optimization towards answer selection. 
% The objective is to learn the optimal $\omega$ that can achieve the highest possible accuracy. 
The weight functions are designed as follows:
\begin{equation*}
    % \small
    \omega_v(v^{c_i}_i; \theta) = \begin{cases}
        \theta_{gr} R(v^{c_i}_i) &  v^{c_i}_i \in F^{c_i}_{G}\\
        \theta_{ar} R(v^{c_i}_i) &  v^{c_i}_i \in F^{c_i}_{A}\\
        0 & v^{c_i}_i = \{Q||c_i\} \\
        \end{cases}
\end{equation*}
\begin{equation*}
\resizebox{\columnwidth}{!}{%
$\omega_e(v^{c_i}_j,v^{c_i}_k;\theta) =
\begin{cases}
\theta_{gg}O_(v^{c_i}_j,v^{c_i}_k) & v^{c_i}_j,v^{c_i}_k\in F^{c_i}_{G}\\
\theta_{aa} O(v^{c_i}_j,v^{c_i}_k) & v^{c_i}_j,v^{c_i}_k \in F^{c_i}_{A}\\
\theta_{ga}\ceil*{O(v^{c_i}_j,v^{c_i}_k)}  & v^{c_i}_j \in F^{c_i}_{G} ,v^{c_i}_k \in F^{c_i}_{A}\\
\theta_{qg}\ceil*{O(v^{c_i}_j,v^{c_i}_k)}  & v^{c_i}_j \in F^{c_i}_{G} ,v^{c_i}_k = {Q||c_i}\\
\theta_{qa}O(v^{c_i}_j,v^{c_i}_k)  & v^{c_i}_j \in F^{c_i}_{A} ,v^{c_i}_k = {Q||c_i} \\
\end{cases}$
}
\end{equation*}

As evident from the equation, the ceiling operation ($\ceil*{O}$) is applied to the overlap score for question-grounding and grounding-abstract edges. While other edges are concerned about how many terms are shared within the nodes, grounding-abstract and question-grounding are expected to cover only one term.

The LP solver will seek to extract the optimal subgraph with the highest possible sum of node and edge weights. Given this premise, we model the overlap we want to minimize with a negative coefficient and the overlaps we want to maximize with a positive coefficient. Given this premise, the domains of parameters are imposed as: $\theta_{vw},~\theta_{ew},~\theta_{gr},~\theta_{ar},~\theta_{aa},~\theta_{ga},~\theta_{qg},~\theta_{qa} \in [0.0,1.0]$ and $\theta_{gg} \in [-1.0,0.0]$
% \begin{align*}
%     \theta_{gr},~\theta_{ar},~\theta_{aa},~\theta_{ga},~\theta_{gq},~\theta_{aq} &\in [0.0,1.0]\\
%     \theta_{gg}&\in [-1.0,0.0]
% \end{align*}

We use the following binary variables to represent the presence of nodes and edges in the subgraph:

\noindent \textbf{(1)} Binary variable $y_v$ takes the value of 1 iff $v \in V^{c_i}$ belongs to the subgraph.\\
\noindent \textbf{(2)} Binary variable $z_e$ takes the value of 1 iff $e \in E^{c_i}$ belongs to the subgraph.

In order to emulate the grounding-abstract inference chains and obtain a valid subgraph, we impose the set constraints stipulated in Table~\ref{tab:lp_constraints} for the LP solver.

\noindent\textbf{Answer Selection}: Given Question $Q$ and choices $C=\{c_1,~c_2,~c_3, ...,~c_n\}$ we extract optimal explanation graphs $G^Q=\{\hat{G}^{c_1},~\hat{G}^{c_2},~\hat{G}^{c_3},...,~\hat{G}^{c_n}\}$ for each choices. We hypothesize that the graph built from the correct answer will have the highest node and edge weights. Based on this premise we define the correct answer $c_{ans}$ as $c_{ans} = \argmax_{c_i}~(\Omega(\hat{G}^{c_i}))$.

\noindent\textbf{Bayesian Optimization}: Unlike neural approaches, our approach's gradient function is intractable. In order to learn the optimal values for the parameters, we surrogate the accuracy function with a multi-variate Gaussian distribution $\mathcal{N}_{9}(\mu,\,\sigma^{2})\,$ and maximize the accuracy by performing Bayesian optimization. To the best of our knowledge, our system is the first to employ Bayesian optimization with linear programming to solve question answering.
% \vspace{-\baselineskip}
\section{Empirical Evaluation}

\paragraph{Background Knowledge for Science QA}: As stated in Section~\ref{sec:approach}, ExplanationLP requires two different knowledge bases: \textit{Abstract KB} and \textit{Grounding KB}. We construct the required knowledge bases using the following sources.

\noindent\textbf{(1) Abstract KB}: Our Abstract knowledge base is constructed from the WorldTree Tablestore corpus~\cite{jansen2018worldtree}. The Tablestore corpus contains 4,097 common sense and scientific facts to create explanations for multiple-choice science questions. The corpus is built for answering elementary science questions encouraging possible knowledge reuse to elicit explanatory patterns. We extract the core scientific facts to build the Abstract KB. Core scientific facts are independent of a specific question and represent the central concept a question is testing, such as {\texttt{Actions}} (\textit{friction occurs when two object’s surfaces move against each other}) or {\texttt{Affordances}} (\textit{friction causes the temperature of an object to increase}). To this end, we select the facts that play a \textit{central role} in the gold explanations.

\noindent\textbf{(2) Grounding KB}: The grounding knowledge base should consist of definitional knowledge (e.g., synonymy and taxonomy)  that can take into account different variations of questions and help it link it to abstract facts. To achieve this goal, we select the \textit{is-a} and \textit{synonymy} facts from ConceptNet~\cite{speer2016conceptnet} as our grounding facts. ConceptNet has higher coverage and precision, enabling us to answer a wide variety of questions. \\

\noindent\textbf{Question Sets}:  We use the following question sets to evaluate ExplanationLP's performance and compare it against other explainable approaches. 

\noindent\textbf{(1) WorldTree Corpus}: The 2,290 questions in the WorldTree corpus are split into three different subsets: \emph{train-set} (987), \emph{dev-set} (226) and \emph{test-set} (1,077). We use the \emph{dev-set} to assess the explainability performance and error analysis since the explanations for \emph{test-set} are not publicly available.

\noindent\textbf{(2) ARC-Challenge Corpus}: ARC-Challenge is a multiple-choice question dataset which consists of question from science exams from grade 3 to grade 9~\cite{clark2018think}. We only consider the Challenge set of questions. These questions have proven to be challenging to answer for other LP-based question answering and neural approaches. ExplanationLP rely only on the \emph{train-set} (1,119) and test on the \emph{test-set} (1,172). ExplanationLP does not require \emph{dev-set}, since the possibility of over-fitting is non-existent with only nine parameters.\\

\noindent\textbf{Relevant Facts Retrieval (FR)}:
We experiment with two different fact retrieval scores. The first model -- i.e. \emph{BM25 Retrieval}, adopts a BM25 vector  representation for question, candidate answers and explanation facts. We apply this retrieval for both Grounding and Abstract retrieval. We use the IDF score from BM25 as our downstream model's relevance score. The second approach -- i.e. \emph{Unification Retrieval (UR)}, represents the BM25 implementation of the Unification-based Reconstruction framework described in \citet{valentino2020unification}. The unification score for a given fact depends on how often the same fact appears in explanations for similar questions. Considering that the unification retrieval relies on gold explanations, we only employ this approach to retrieve abstract facts and BM25 retrieval for grounding facts. Unification retrieval has shown to be scalable and exhibits better performance for retrieving abstract explanatory facts than transformer-based approaches on the WorldTree explanation reconstruction task.\\
% The framework combines the BM25 relevance score with a unification score, representing the fact's explanatory power. 
 
% For unification retrieval, we use the Unification score as our downstream model's relevance score. 
%while the similarity function between the questions is realized through BM25 vectors and cosine similarity. 
% The Unification Retrieval model has shown better performance when compared to the BM25 Retrieval model in the task of ranking explanatory sentences for multiple-choice science questions.
% , improving the retrieval of complex scientific facts that have low degree of overlaps with question and candidate answer.  

\noindent\textbf{Baselines}: The following baselines are replicated on the WorldTree corpus to compare against ExplanationLP: 

\noindent \textbf{(1) Bert-Based models}: We compare the ExplanationLP model's performance against a set of BERT baselines. The first baseline -- i.e. \emph{BERT$_{Base}$}/\emph{BERT$_{Large}$}, is represented by a standard BERT language model~\cite{devlin2019bert} fine-tuned for multiple-choice question answering. Specifically, the model is trained for binary classification on each question-candidate answer pair to maximize the correct choice (i.e., predict 1) and minimize the wrong choices (i.e., predict 0). During inference, we select the choice with the highest prediction score as the correct answer. BERT baselines are further enhanced with explanatory facts retrieved by the retrieval models. The first model, \emph{BERT + BM25}, is fine-tuned for binary classification by complementing the question-answer pair with grounding and abstract facts selected by BM25 retrieval. Similarly, the second model \emph{BERT + UR} complements the question-answer pair with grounding and abstract facts selected using BM25 and Unification retrieval, respectively.

\noindent \textbf{(2) PathNet}~\cite{kundu2018exploiting}: PathNet is a neural approach that constructs a single linear path composed of two facts connected via entity pairs for reasoning. PathNet also can explain its reasoning via explicit reasoning paths. They have exhibited strong performance for multiple-choice science questions while adopting a two-hop reasoning strategy. Similar to  BERT-based models, we employ PathNET with the top $k$ facts retrieved utilizing Unification (\emph{PathNet + UR}) and BM25 (\emph{PathNet + BM25}) retrieval. We concatenate the facts retrieved for each candidate answer and provide as supporting facts.

% Further details regarding the hyperparameters and code used for each model, along with information concerning the knowledge base construction and dataset information, can be found in the Technical Appendix.

\subsection{Answer Selection}

\setlength{\tabcolsep}{3pt}

\begin{table}[!htb]
    \centering
    \small
    \begin{tabular}{@{}cp{3.5cm}cccc@{}}
    \toprule
    \# & \textbf{Model}  & \multicolumn{4}{c}{\textbf{Accuracy}}  \\
    && \textit{k}=20 & \textit{k}=30 & \textit{k}=40 & \textit{k}=50 \\
    %  \midrule
    % \multicolumn{6}{c}{\textbf{Baselines}} \\
    \midrule
    1 & BERT$_{Base}$  & \multicolumn{4}{c}{44.56}\\
    2 & BERT$_{Large}$ & \multicolumn{4}{c}{47.26} \\
    \midrule
     3 & BERT$_{Base}$ + BM25   & 43.63 &37.97 & 31.56 & 32.68   \\
    4 &  BERT$_{Base}$ + UR & 42.14 & 31.84 & 30.36 & 31.29  \\
     5 & BERT$_{Large}$ + BM25 & 43.36 & 32.86 & 35.46 & 26.92 \\
    6 &  BERT$_{Large}$ + UR  & 42.98 & 27.39 & 24.88 & 26.55   \\
     \midrule
      7 & PathNet + BM25  & 41.61 & 41.98 &40.11 & 41.79 \\ 
      8 & PathNet + UR & 43.58 & 40.76 & 41.22& 42.83  \\ 
      \midrule
    % \multicolumn{6}{c}{\textbf{ExplanationLP}} \\
    \midrule
    9 & ExplanationLP + BM25  & 50.88 & 50.51 & 50.78 & 50.41  \\ 
    10 & ExplanationLP + UR  &  55.24 &  \underline{\textbf{56.36}} & 54.21 & 54.24 \\ 
     \bottomrule
    \end{tabular}
    \caption{\small Answer selection performance on the WorldTree \emph{test-set}. \textit{k} represents the number of retrieved facts by the respective retrieval approaches.}
    \label{tab:wt_answer_selection_k}
    \centering
    \begin{tabular}{@{}cp{5cm}ccc@{}}
    \toprule
    \# & \textbf{Model} & \multicolumn{2}{c}{\textbf{Accuracy}}  \\
    &&  \textbf{Easy} & \textbf{Challenge} \\
    % \midrule
    % \multicolumn{4}{c}{\textbf{Baselines}} \\
    \midrule
    11 & BERT$_{Base}$ & 51.04 & 28.75  \\
    12 & BERT$_{Large}$ & 54.58 & 29.39 \\
    \midrule
     13 & BERT$_{Base}$ + BM25 (\textit{k}=20) & 45.54 & 38.97       \\
    14 &  BERT$_{Base}$ + UR (\textit{k}=20)& 44.10 & 37.80  \\
     15 & BERT$_{Large}$ + BM25 (\textit{k}=20) & 45.94 & 37.06   \\
    16 &  BERT$_{Large}$ + UR (\textit{k}=20) & 43.97 & \underline{\textbf{40.57}}    \\
     \midrule
     17 & PathNet + BM25 (\textit{k}=20)  & 43.32 & 36.42   \\ 
      18 & PathNet + UR (\textit{k}=15) & 47.64 & 33.55   \\ 
     \midrule
    % \multicolumn{4}{c}{\textbf{ExplanationLP}} \\
    \midrule
    19 & ExplanationLP + BM25 (\textit{k}=20) & 54.49 & 39.61 \\ 
    20 & ExplanationLP + UR (\textit{k=}20) & \underline{\textbf{62.82}} & \underline{\textbf{40.57}} \\ 
     \bottomrule
    \end{tabular}
    \caption{\small Accuracy on Easy (764) and Challenge split (313) of WorldTree \emph{test-set} corpus from the best performing \textit{k} of each model}
    \label{tab:wt_answer_selection_split}
\end{table}

\begin{table*}[!htb]
    \centering
    \small
    \begin{tabular}{@{}cp{9cm}ccP{2cm}c@{}}
    \toprule
    \# & \textbf{Model} & \textbf{Explainable} &\textbf{Pre-Trained} & \textbf{External KB} & \textbf{Accuracy}  \\
    % \midrule
    % \multicolumn{6}{c}{\textbf{Previous Works (from ARC Leaderboard)}} \\
    \midrule
     1 & TupleInf~\cite{Khot_2017} & Yes & No  & Yes&23.83   \\ 
     2 & TableILP~\cite{khashabi2016question} & Yes & No  & Yes&  27.11  \\ 
     3 & IR-Solver~\cite{clark2016combining} & Yes & No  & No &20.26  \\ 
     4 & DGEM~\cite{khot2018scitail} & Partial &Yes  & Yes &  27.11 \\

     5 & KG$^2$~\cite{zhang2018kg} & Partial  & Yes & Yes &  31.70 \\
    %  \midrule
     6 & AHE (FLAIR+BERT)~\cite{yadav2019alignment} & Partial & Yes & No  &33.87 \\ 
     7 & AHE (FLAIR+BERT+GloVe+InferSent) with grade~\cite{yadav2019alignment} & Partial & Yes & No & \textbf{34.47} \\ 
    % \midrule
    % \multicolumn{6}{c}{\textbf{Baselines}} \\
    % \midrule
    % 5 & BERT$_{Base}$ & N &   50.00 \\
    % 6 & BERT$_{Large}$ & N &  50.00 \\
    % \midrule
    %  7 & BERT$_{Base}$ + BM25 (\textit{k}=) &P &  36.07   \\
    %  8 & BERT$_{Large}$ + BM25 (\textit{k}=25) & P &  43.45  \\
    %  \midrule
    %  9 & PathNet + BM25 (\textit{k}=)  &Y&  65.18   \\ 
    
     \midrule
    % \multicolumn{6}{c}{\textbf{ExplanationLP}} \\
    \midrule
    8 & ExplanationLP + BM25 (\textit{k}=20) & Yes & No & Yes &\textbf{32.16}  \\ 
    9 & ExplanationLP + UR (\textit{k}=30) & Yes & No & Yes & 30.54  \\ 
     \bottomrule
    \end{tabular}
    \caption{\small ARC challenge scores compared with other Fully or Partially explainable approaches.}
    \label{tab:arc_answer_selection}
\end{table*}

\noindent\textbf{WorldTree Corpus}: Table~\ref{tab:wt_answer_selection_k} reports the analyses carried out with BERT, PathNet and ExplanationLP on the WorldTree \emph{test-set} for varying top \textit{k} relevant facts. We also report the score on the Easy and Challenge subsets for each model's best performance setting in Table~\ref{tab:wt_answer_selection_split}. We retrieve top $l$  relevant grounding facts from Grounding KB and top $m$ relevant abstract facts from Abstract KB such that $l + m = k$ and $l = m$. To ensure fairness across the approaches, the same amount of facts are presented to each model. A range of \textit{k} values are tested to evaluate the robustness of approaches with increasing distracting information.

\noindent Given the above premises, the following conclusions can be drawn from  observing the scores in Table~\ref{tab:wt_answer_selection_k} and Table~\ref{tab:wt_answer_selection_split}:

\noindent (1) Despite having a smaller number of parameters to train (BERT$_{Base}$: 110M parameters, BERT$_{Large}$: 340M parameters, ExplanationLP: 9 parameters), the best performing ExplanationLP (\#10, \textit{k}=30) overall outperforms both BERT$_{Base}$ (\#1) and BERT$_{Large}$ (\#2) with no explanations by 11.8\% and  9.1\% respectively. ExplanationLP also outperforms BERT with no explanations in both Easy and Challenge split by 8.28 (\#12,\#20) and 11.18 (\#12,\#20), respectively.

\noindent (2) The lowest drop in performance of ExplanationLP with BM25 is 0.46\% (\#9, \textit{k}=20 $\rightarrow$ \textit{k}=50) and UR is  2.42\% (\#10, \textit{k}=30 $\rightarrow$ \textit{k}=50). On the other hand, the lowest drop in performance for BERT$_{Large}$ with BM25 is 16.44\% (\#5, \textit{k}=20, $\rightarrow$ \textit{k}=50) and UR is  16.43\% (\#6, \textit{k}=20 $\rightarrow$ \textit{k}=50). Similarly, BERT$_{Base}$ with BM25 and UR has the lowest drop of performance of 10.95\% and 10.85\%, respectively. These scores along with overall steady drop in performance  with increasing \textit{k} indicates that BERT struggles with the increasing number distracting knowledge. In contrast, ExplanationLP can operate on a higher amount of distracting information and still obtain better scores, displaying resilience towards noise. It should also be noted that BERT$_{Base}$ outperforms BERT$_{Large}$ for BERT + FR because BERT$_{Base}$ is small enough to faciliate a larger batch size resulting in a better learning. 

\noindent (3) BERT, by itself, does not provide any explanations. BERT + Fact Retrieval (FR) is comparatively better since the facts provided by the FR could function as an explanation. However, BERT is still inherently a black-box model, not being entirely possible to explain its prediction. By contrast, ExplanationLP is fully explainable and produces a complete explanatory graph. The model design, by principle, supports full interpretability of the reasoning behind the answer prediction.

\noindent (4) Similar to ExplanationLP, PathNet is also explainable and demonstrates robustness to noise. However, we outperform PathNet consistently across different retrievals and \textit{k} values. ExplanationLP also outperforms PathNet's best performance setting by 12.78\% (\#10, \textit{k}=20 and \#8, \textit{k}=20). We also demonstrate significantly better scores on both Easy and Challenge split.

\noindent (5) ExplanationLP consistently exhibits better scores on both BM25 and UR compared to BERT and PathNet, demonstrating independence of the upstream retrieval model for performance.

\noindent\textbf{ARC-Challenge}: We select the best performance setting on the WorldTree corpus and experiment on the ARC-Challenge corpus~ \cite{clark2018think}  to evaluate ExplanationLP on a more extensive general question set and compare against SOTA approaches that provide explanations for inference. Table \ref{tab:arc_answer_selection} reports the results on the \emph{test-set}. We compare ExplanationLP against published approaches along the following dimensions: 
(1) \emph{Explainable}: Indicates if the model produces an explanation/evidence for the predicted answer. A subset of the approaches produces evidence for the answer but remains intrinsically black-box. These models have been marked as \textit{Partial}; (2) \emph{Pre-trained}: Denotes if the approach uses pre-trained models like Language models and/or Word Embeddings; (3) \emph{External KB}: Indicates if the system uses external knowledge bases other than the one provided by the ARC-Challenge dataset.

The results show that ExplantionLP outperforms other fully explainable and information retrieval based baselines. We also exceed specific neural approaches that provide evidence for inference. At the same time, ExplanationLP also exhibits competitive performance with black-box models that have been pre-trained on external datasets and supervised with a large number of parameters without any pre-training and fewer number of parameters.

\subsection{Explanation Retrieval}
\label{sec:explanation_retrieval}

\begin{table}[!htb]
    \small
    \centering
\begin{tabular}{lcccc}
    \toprule
    \textbf{Approach}  & \multicolumn{4}{c}{\textbf{F1@\textit{k}}}\\
    \cmidrule{2-5}
    & \textit{k=}20 & \textit{k=}30 & \textit{k=}40 & \textit{k=}50 \\
    \midrule
      UR Only & 33.08 & 24.63 & 21.08 & 19.03  \\
      UR + ExplanationLP & 44.00 & 44.07 & 43.88 & 43.88 \\
      \midrule
      BM25 Only  & 24.62 & 20.23  &17.16 & 15.19\\
      BM25 + ExplanationLP  &  36.01 &35.93&35.81&35.81\\
    \bottomrule
    \end{tabular}
    \caption{\small Explanation retrieval performance on the WorldTree Corpus \emph{dev-set}. \textit{k} indicates the number of retrieved facts by the respective retrieval approaches.  }
    \label{tab:explanation_selection}
\end{table}

% \begin{figure*}
%     \centering
%     \includegraphics[width=\textwidth]{images/correct_answer.pdf}
%     \caption{An example of explanations and scores retrieved from ExplanationLP. The answer for this question is \textit{first quarter} which was correctly predicted.}
%     \label{fig:correct_answer}
% \end{figure*}

Table~\ref{tab:explanation_selection} shows the F1 score for explanation retrieval (abstract facts only) of varying number of facts  (\textit{k} facts) retrieved using the FR approaches. As the number of facts increases, the F1 score of FR approaches drops drastically, indicating the increase in distracting knowledge. This stability in performance with the increasing number of distractors adds evidence to ExplanationLP's robustness towards noise and explainability.

% Figure~\ref{fig:correct_answer} illustrates a qualitative example of depicts a set of explanations and scores retrieved by our approach. We can observe for the wrong choices $F_{G2}$ is chosen as a grounding because of the word \textit{picture} in the question (Question-Grounding relevance). This fact, in turn, links to $F_{C1}$ (Grounding-Central inter-hop overlap). For the correct choice, $F_{G5}$ links to a different abstract fact which shares lexical clustered near $F_{C4}$ (Central-Central intra-hop overlap). $F_{C1}$ does not share a lexical overlap with any of the selected abstract facts; hence ExplanationLP avoids selecting this and consequently not select $F_{G2}$ (Abstract fact limit constraint). 

\subsection{Interpreting Parameter Scores}
\label{sec:ablation}

\begin{table}[h]
    \small
    \centering
    \resizebox{\columnwidth}{!}{\begin{tabular}{cp{5.5cm}ccc}
    \toprule
    \textbf{\#} & \textbf{Parameter} & \multicolumn{2}{c}{\textbf{Value}} \\
    &&\textbf{WT} & \textbf{ARC}\\
    \midrule
     1 & Question-Abstract overlap ($\theta_{qa}$) & 0.10 & 0.09 \\
     2 & Question-Grounding overlap ($\theta_{qg}$) & 0.98 & 0.84 \\
     3 & Abstract-Abstract overlap ($\theta_{aa}$) & 0.01 & 0.11 \\
     4 &  Grounding-Abstract overlap ($\theta_{ga}$) & 0.14 & 0.23   \\
     5 & Grounding-Grounding overlap  ($\theta_{gg}$) & -0.99 & -0.92 \\
     \midrule
     6 & Abstract Relevance ($\theta_{ar}$) & 0.03 & 0.09  \\
     7 & Grounding Relevance ($\theta_{gr}$) & 0.36 & 0.14 \\
     \midrule
    %  8 &  Coverage weight ($\theta_{cw}$) & 0.54 & 0.23  \\
     8 &  Edge weight ($\theta_{ew}$) & 0.80 & 0.26 \\
     9 & Node weight ($\theta_{vw}$) & 0.76 & 0.67 \\
    \bottomrule
    \end{tabular}}
    \caption{Parameter scores obtained from optimizing  ExplanationLP (best performance setting) for answer selection for ARC and WorldTreeCorpus (WT) and ARC-Challenge (ARC) corpus.}
    \label{tab:hyper}
\end{table}

Table~\ref{tab:hyper} reports the parameter values obtained using Bayesian optimization for the best accuracy score. Unlike black-box neural approaches, ExplanationLP's parameters can be interpreted and we draw the following observations from the table:
% validating our research hypotheses (\textbf{RH}) listed in Section~\ref{sec:construct_graph}:

% \noindent (1)  Removing the categorization of grounding-abstract facts lowers the performance by 30.74\% (Row~\#2), showing the impact of grounding-abstract chains in answer selection. 

% \noindent (2)  The difference in performance at Row~\#1 and~\#4 (-30.18\%) validates the importance of relevance scores. The high hyperparameter scores obtained for abstract relevance (Row~\#13) and grounding relevance (Row~\#14) corroborate this finding.

% \noindent (3)  The performance drop between Row~\#1 and~\#5 (-15.33\%) highlights the importance of chaining constraints. Without the chaining constraint, the LP approach would pick the highest relevance and edges with positive values, ignoring negative edges. If negative edges are not selected, the LP system would not be encouraged to have diverse grounding facts. This observation is also consistent with the grounding-grounding overlap hyperparameter score (Row~\#12).
\noindent (1) The optimal score we obtain for question-grounding overlap (\#2) grounding-abstract (\#4) overlap and grounding relevance  (\#7) scores are relatively high. These scores indicate the importance of grounding fact in the final inference.

\noindent (2) The high negative scores the system obtains for grounding-grounding (\#5) overlap validates our assumption that penalizing grounding overlaps would incentivize the system to address different parts of a question and reinforces the importance of grounding-abstract inference chains.

\noindent (3) The abstract-abstract overlap score (\#3), grounding-abstract overlap score (\#1) and abstract relevance (\#6) is relatively smaller, indicating a lower level of importance. However, it still plays a role in the final score, stipulating the need for encouraging conceptual similarity between abstract facts. 

\noindent (4) Most of the scores are consistent over different datasets except node and edge weights (\#8 and \#9). The knowledge bases we use are sufficient enough to answer all the questions in the WorldTree corpus; hence equal importance is given to node and edge weights. In contrast, for the ARC corpus, ExplanationLP relies more on relevance scores than edge weights. 
% These scores can indicate how ExplanationLP adapts to more general question sets.

\subsection{Error Analysis}
\label{sec:error_analysis}
\begin{figure}[t!]
    \centering
    \begin{subfigure}[t]{\columnwidth}
        \centering
        \includegraphics[width=\columnwidth,height=4.2cm]{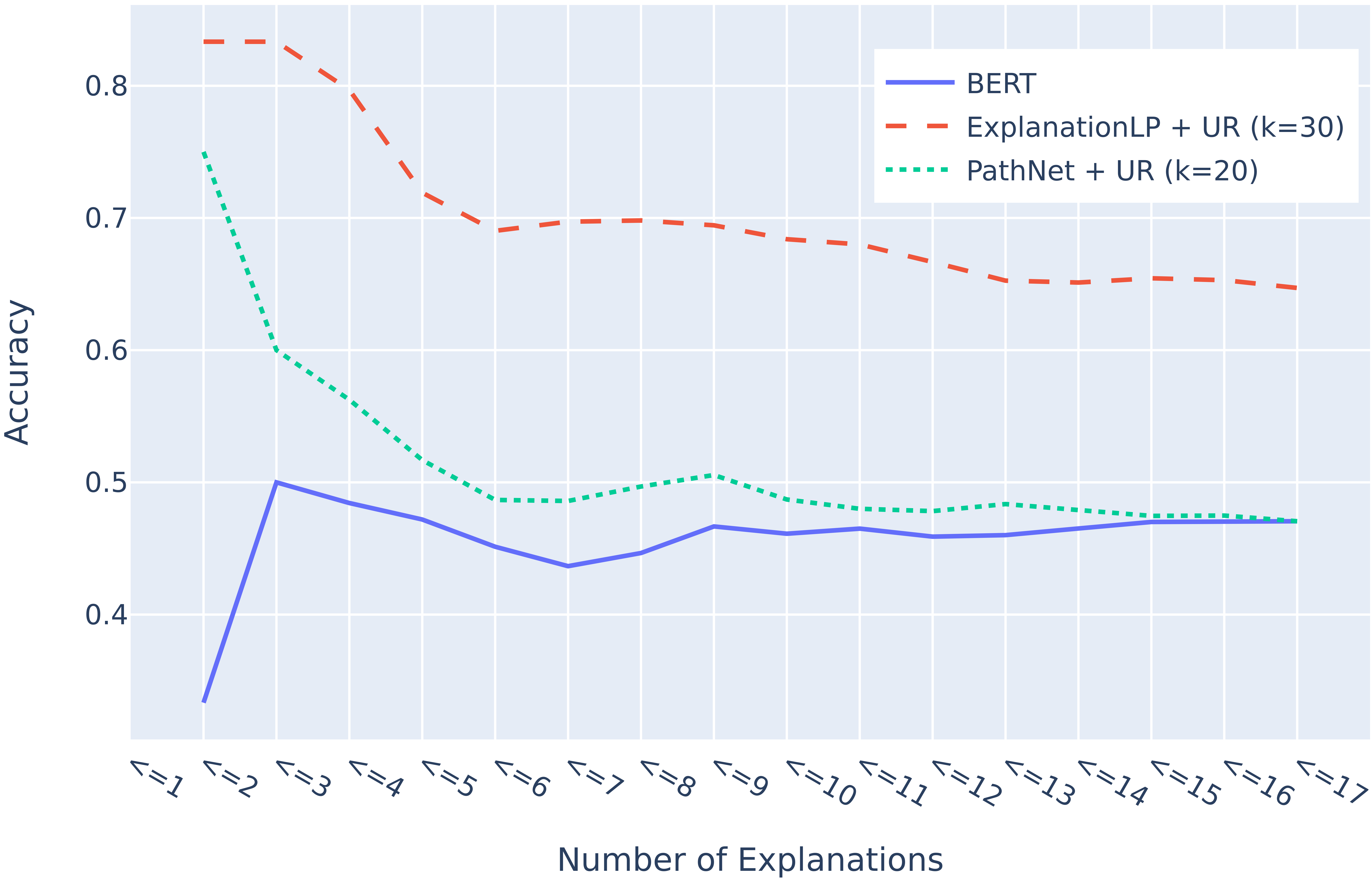}
        \caption{\small Explanation length Vs Accuracy}
        \label{fig:explanation_length}
    \end{subfigure}
    ~ 
    \begin{subfigure}[t]{\columnwidth}
        \centering
        \includegraphics[width=\columnwidth,height=4.2cm]{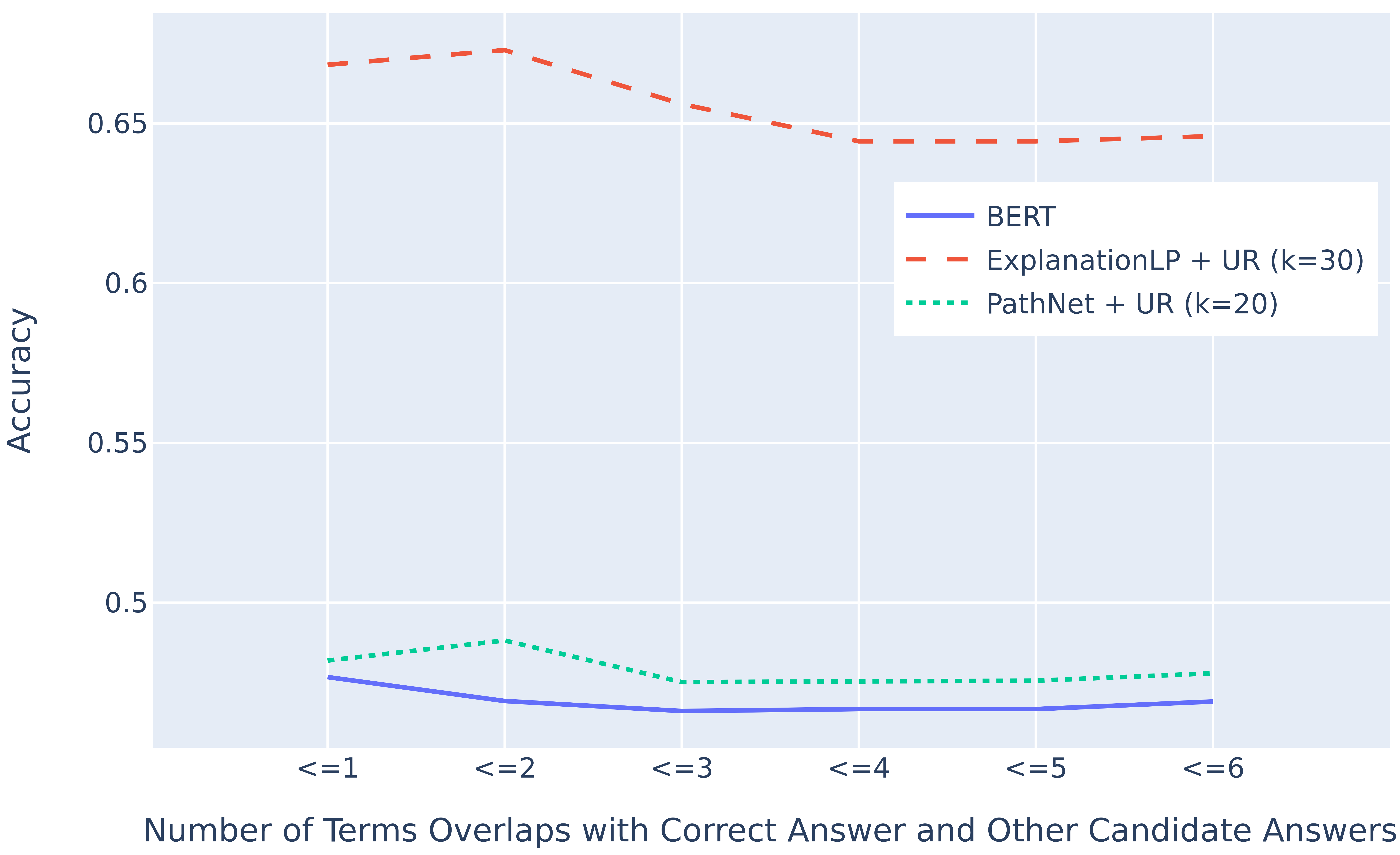}
        \caption{\small Candidate answers term overlaps Vs Accuracy}
        \label{fig:choice_overlap}
    \end{subfigure}
     ~ 
    \begin{subfigure}[t]{\columnwidth}
        \centering
        \includegraphics[width=\columnwidth,height=4.2cm]{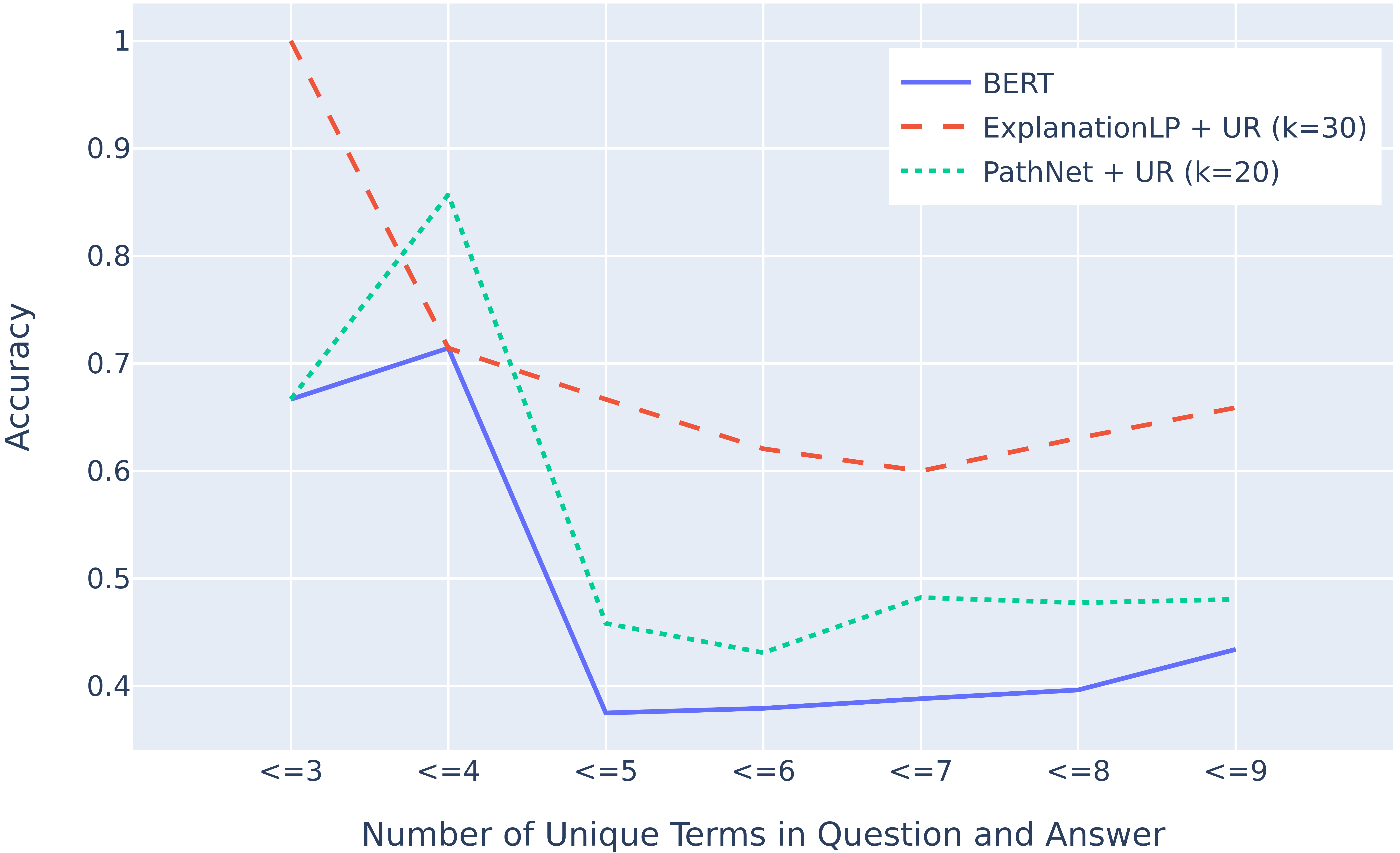}
        \caption{\small Number of unique terms in question and answer Vs Accuracy}
        \label{fig:question_word_count}
    \end{subfigure}
    \caption{\small Quantitative analysis of answer prediction accuracy on the development set varying across different models for WorldTree \emph{dev-set}. Red line represents ExplanationLP + UR (\textit{k}=20), blue line represents BERT$_{Large}$ and  green line represents PathNet + UR (\textit{k}=20)}
\end{figure}

% We present the error analysis for question answering accuracy.

\noindent\textbf{Explanation length.} Figure~\ref{fig:explanation_length} plots the change in accuracy as we consider questions with longer explanations. As demonstrated in explanation regeneration~\cite{valentino2020unification,jansen2019textgraphs}, the complexity of a science question is directly correlated with the explanation length -- i.e. the number of facts required in the gold explanation. Since BERT does not rely on external knowledge, the performance is stable across the spectrum. On the other, both PathNet and ExplanationLP use external background knowledge, addressing the multi-hop process in two main reasoning steps. However, unlike ExplanationLP, PathNet combines only two explanatory facts to answer a given question. This assumption has a negative impact on answering complex questions requiring long explanations. This is evident in the graph, where we observe a sharp decrease in accuracy with increasing explanation length. Comparatively, ExplanationLP achieves more stable performance, showing a lower degradation with an increasing number of explanation sentences. These results crucially demonstrate the positive impact of grounding-abstract mechanisms on semantic drift.

\noindent\textbf{Term overlap between choices.} Figure~\ref{fig:choice_overlap} illustrates the change in accuracy as the term overlap between answer and other candidate answers increases. Despite our approach being dependent on lexical overlaps, there is minimal degradation in performance, indicating that the constraints and structure help alleviate the noise rising from the lexical space. 

\noindent\textbf{Unique terms in question and answer } Figure~\ref{fig:question_word_count} presents the change in accuracy as the number of unique terms increases in question and answer. With an increasing number of terms, there are more distractors; thus, a higher degradation in performance than PathNet and BERT. We believe that rather than treating all terms equally, the performance could be increased by extracting the essential terms.

\section{Related Work}

TableILP~\cite{khashabi2016question} is one of the first approaches employing linear programming for science question answering. TableILP performs reasoning over a semi-structured knowledge base by extracting a subgraph connecting question to the answer. TableILP is followed by TupleILP~\cite{khot2017answering}. Similar to TableILP, TupleILP present XSQA as a graph connection problem. However, in contrast to TableILP, TupleILP performed reasoning over tuples extracted using open information extraction. SemanticILP~\cite{semanticilp2018aaai} is built on top of the same path traversal problem but with multiple semantic abstractions. These approaches treat all facts homogeneously and attempt to alleviate semantic drift by enforcing semantic constraints.
In contrast, our approach explicitly groups facts into grounding and abstract. While there are XSQA approaches that employ neural approaches~\cite{sharp2017tell,jansen2017framing}, our shows that linear programming allows the freedom to encode our inference assumptions. To the best of our knowledge, this is the first approach for solving explainable question answering that attempts to model grounding-abstract inference chains. Our approach also leverages Bayesian optimization with linear programming.

The construction of explanations using graph-based models and explicit inference chains is an active area of research for explainability in question answering and machine reading comprehension~\cite{thayaparan2020survey}. The PathNet approach used in the empirical evaluation extracts explicit inference chains for science question answering connecting two facts from an external corpus~\cite{kundu2018exploiting}. Systems that apply graph neural networks for structured reasoning have also demonstrated competitive performances for multi-hop open domain question answering~\cite{tu2020select,fang2019hierarchical,thayaparan2019identifying}.  Graph-based approaches have also been employed for mathematical reasoning~\cite{ferreira2020natural,ferreira2020premise} and textual entailment~\cite{silva2018recognizing,silva2019exploring}. Other methods construct explanation chains by leveraging explanatory patterns emerging in a corpus of scientific explanations \cite{valentino2020explainable,valentino2020unification}.

% ~\citet{khashabi2019capabilities} proposed a theoretical framework showing how long chains of inference are not feasible. These findings are also empirically supported by~\cite{jansen2018worldtree,fried2015higher}. 

% \cite{sharp2017tell} - Uses neural networks to learn the justification scores max pool them to use as candidate answer score. Uses lexical features and discourse representation features to learn the scores. 

% \cite{jansen2017framing} similar to previous. Builds lexical overlap intersentence and external knowledge. This lexical overlap is identified using semantic information. 

% \cite{yadav2019quick} - Quick not so dirty - Scoring functions. Functions as justifications.

\section{Conclusion}

In this paper, we presented ExplanationLP, an explainable science question answering model that performs abductive reasoning on grounding-abstract chains. The paper presented an in-depth systematic evaluation over a comprehensive set of semantic features, outlining the impact of various research assumptions. We also showed how ExplanationLP  can effectively generalize for general science question sets.  Despite the challenge of operating on noisy lexical space, we achieve SOTA scores for Explainable Science QA, attesting to ExplanationLP's ability to build better inference chains.
% We have empirically and conceptually demonstrated the advantage of explicitly modeling grounding and abstract mechanisms while showing Bayesian optimization with linear programming for QA.

% \FloatBarrier
\bibliography{main}
\bibliographystyle{acl_natbib}

\end{document}